\documentclass{article}
\usepackage{spconf}

\usepackage{cite}
\usepackage{amsmath,amssymb,amsfonts}
\usepackage{algorithmic}
\usepackage{graphicx}
\usepackage{textcomp}
\usepackage{xcolor}

\usepackage{array}
\usepackage{mathtools}
\usepackage{tcolorbox}
\usepackage{color}
\usepackage{theorem}
\usepackage{amssymb}
\usepackage{caption}
\usepackage{subcaption}
\usepackage{cite,hyperref}
\usepackage{cases}
\usepackage{url}
\usepackage{algorithmic}
\usepackage{algorithm}
\usepackage{tikz}
\usetikzlibrary{shapes,arrows}
\input{my_styles.sty}

\usepackage{multirow}
\usepackage{hhline}

\DeclarePairedDelimiter{\ceil}{\lceil}{\rceil}
\DeclarePairedDelimiter{\floor}{\lfloor}{\rfloor}

\newtheorem{mytheorem}{\bf Theorem}

\newtcolorbox{myblockt}[1]{colback=urblue!5!white,
	colframe=urblue,fonttitle=\bfseries,
	title=#1}

\newtcolorbox{myblock}{colback=urblue!5!white,
	colframe=urblue,fonttitle=\bfseries}

\def\BibTeX{{\rm B\kern-.05em{\sc i\kern-.025em b}\kern-.08em
    T\kern-.1667em\lower.7ex\hbox{E}\kern-.125emX}}

\begin{document}
\ninept

\title{SC-MAD: Mixtures of Higher-order Networks for Data Augmentation}
\name{Madeline Navarro and Santiago Segarra 
\thanks{This work was partially supported by the NSF under award CCF-2008555. 
Research was sponsored by the Army Research Office and was accomplished under Grant Number W911NF-17-S-0002. The views and conclusions contained in this document are those of the authors and should not be interpreted as representing the official policies, either expressed or implied, of the Army Research Office or the U.S. Army or the U.S. Government. The U.S. Government is authorized to reproduce and distribute reprints for Government purposes notwithstanding any copyright notation herein.
Emails:  \href{mailto:nav@rice.edu}{nav@rice.edu}, \href{mailto:segarra@rice.edu}{segarra@rice.edu}}}
\address{Electrical and Computer Engineering, Rice University, USA}
\maketitle

\newcommand \mad[1]        {{\color[cmyk]{1, 0.93, 0.28, 0.22}[Mad: #1]}}
\newcommand \santi[1]        {{\color[cmyk]{0, 0, 0, 0.45}[S: #1]}}

\begin{abstract}
The myriad complex systems with multiway interactions motivate the extension of graph-based pairwise connections to higher-order relations.
In particular, the simplicial complex has inspired generalizations of graph neural networks (GNNs) to simplicial complex-based models.
Learning on such systems requires large amounts of data, which can be expensive or impossible to obtain. 
We propose \emph{data augmentation of simplicial complexes through both linear and nonlinear mixup mechanisms} that return mixtures of existing labeled samples. 
In addition to traditional pairwise mixup, we present a convex clustering mixup approach for a data-driven relationship among several simplicial complexes. 
We theoretically demonstrate that the resultant synthetic simplicial complexes interpolate among existing data with respect to homomorphism densities. 
Our method is demonstrated on both synthetic and real-world datasets for simplicial complex classification.
\end{abstract}

\begin{keywords}
    Simplicial complex, complexon, data augmentation, mixup, convex clustering
\end{keywords}

\section{Introduction}
\label{s:intro}
Simplicial complexes unlock useful topological tools for data science~\cite{barbarossa2020topological,hatcher2002algebraic,schaub2021signal,roddenberry2022hodgelets,roddenberry2019hodgenet} and practical applications~\cite{kanariTopologicalRepresentationBranching2018,romanSimplicialComplexbasedApproach2015} due to their ability to model higher-order interactions.
Simplicial complex-based learning has received much attention lately, with the classical graph-based architectures naturally being extended to higher-order networks~\cite{gohSimplicialAttentionNetworks2022a,ebliSimplicialNeuralNetworks2020,roddenberry2021principled,cinque2023pooling}.
However, graph datasets suffer from limited data due to the complexity of obtaining labeled samples, a problem which is exacerbated for higher-order simplicial complex data.

Data augmentation enables generating synthetic labeled samples from existing data, where the new samples embody characteristics that promote desirable model behavior. 
This procedure is not affected by any machine learning model restrictions as we merely add to the samples present in the dataset, affecting neither model capacity nor the original data~\cite{hernandez2018data,zhang2018mixup}. 
Mixup serves as an efficient data augmentation method that generates new labeled data as mixtures of existing samples~\cite{zhang2018mixup}, and its benefits enjoy copious empirical and theoretical validation~\cite{Thulasidasan2019on,zhangHowDoesMixup2021}.

While graph mixup is still nascent, it has exploded in popularity due to the myriad interesting approaches for interpolating such discrete complex objects~\cite{han2022gmixup,navarroGraphmadGraphMixup2023,wangMixupNodeGraph2021}.
However, data augmentation for higher-order networks is extremely limited~\cite{wei2022augmentations}, and to the best of our knowledge mixup for higher-order networks has never been considered.
Indeed, even in the case of graphs, interpolation of these non-Euclidean objects is nontrivial due to their irregular structure.
This difficulty extends further for simplicial complexes as we must obtain mixtures accounting not only for interconnected entities but also for information shared across dimensions.
Even data augmentation methods amenable to discrete graph objects struggle as higher dimensions are considered~\cite{wei2022augmentations,you2021graph}.
We are thus prompted to turn to the attractive approach of performing mixup in a continuous latent embedding space.
The choice and design of this embedding space allow us to control which characteristics are preserved during the mixup process.

Defining limits of discrete objects enables useful operations for moving within a space of objects as if they are continuous.
The invention of the graphon, the limit object of a convergent sequence of dense graphs, provides a compact continuous space in which the graphs are dense~\cite{lovaszLargeNetworksGraph2012a}.
Graphons allow us to perform tasks on graph data typically restricted to continuous objects, such as barycenter obtention and interpolation for mixup~\cite{xuLearningGraphonsStructured2021a,navarroGraphmadGraphMixup2023,han2022gmixup}.
We can extend this benefit to higher-order networks through the complexon~\cite{roddenberry2023limits}, an analogous limit object for simplicial complexes.

In this work, we present an inaugural method for Simplicial Complex Mixup for Augmenting Data (SC-MAD).
Similarly to existing graph mixup methods~\cite{han2022gmixup,navarroGraphmadGraphMixup2023,wangMixupNodeGraph2021}, we consider a continuous embedding space for the practical implementation of simplicial complex mixup. 
We use the space of complexons, as its being the closure of the space of simplicial complexes means that we can directly compare objects in the original and embedding spaces.
Furthermore, we theoretically show that any continuous interpolant that our method obtains preserves useful structural characteristics~\cite{roddenberry2023limits,lovaszLargeNetworksGraph2012a}. 
In addition to traditional pairwise linear mixup~\cite{zhang2018mixup}, we apply convex clustering for mixup~\cite{pelckmans2005convex,hocking2011clusterpath,lindsten2011clustering}, where new samples describe the mixture of several simplicial complexes~\cite{navarroGraphmadGraphMixup2023}.


\begin{figure*}
\centering
	\begin{minipage}[c]{.3\textwidth}
		\includegraphics[width=\textwidth]{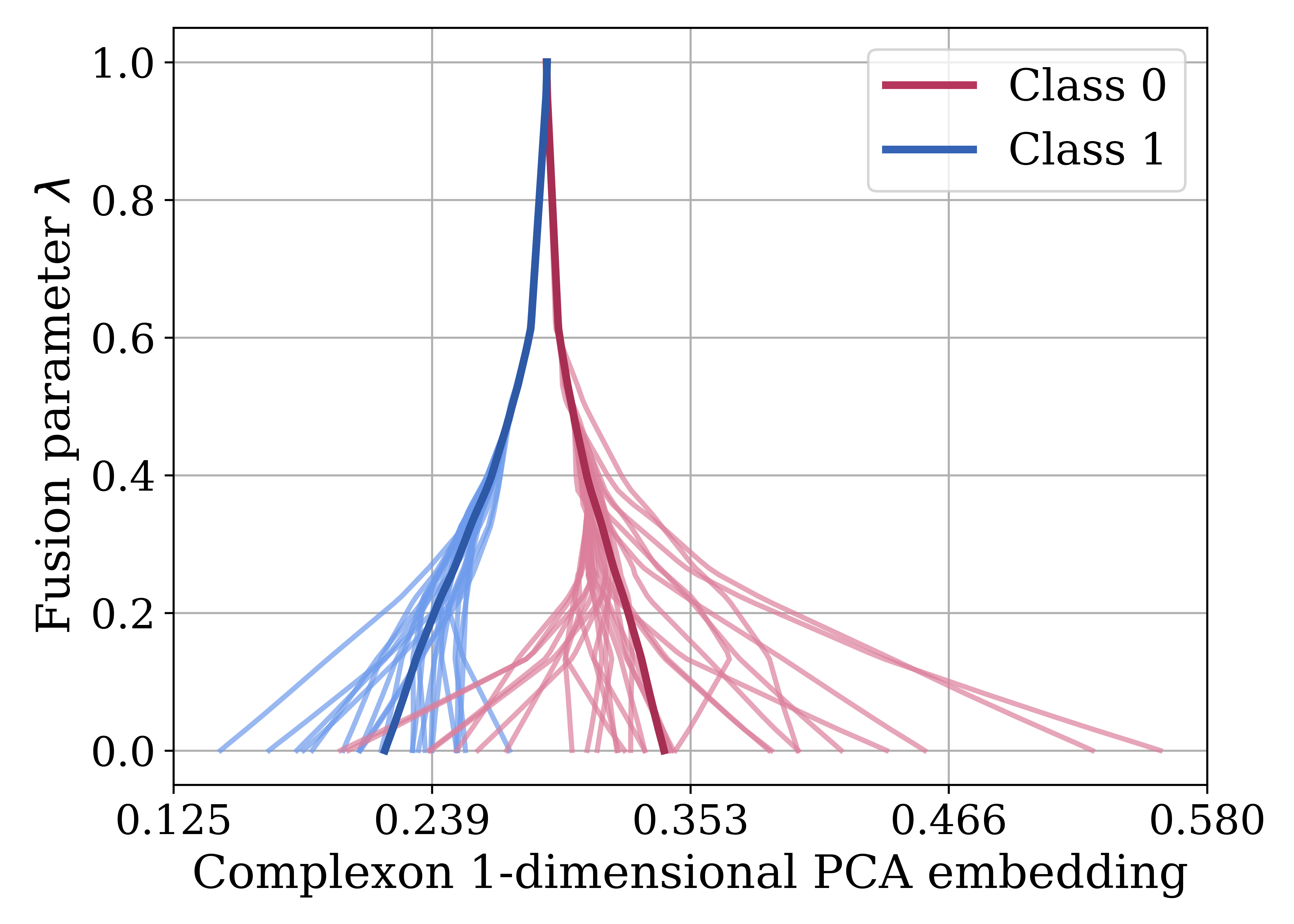}
		\centering{\small (a)}
	\end{minipage}
	\begin{minipage}[c]{.2\textwidth}
		\includegraphics[width=\textwidth]{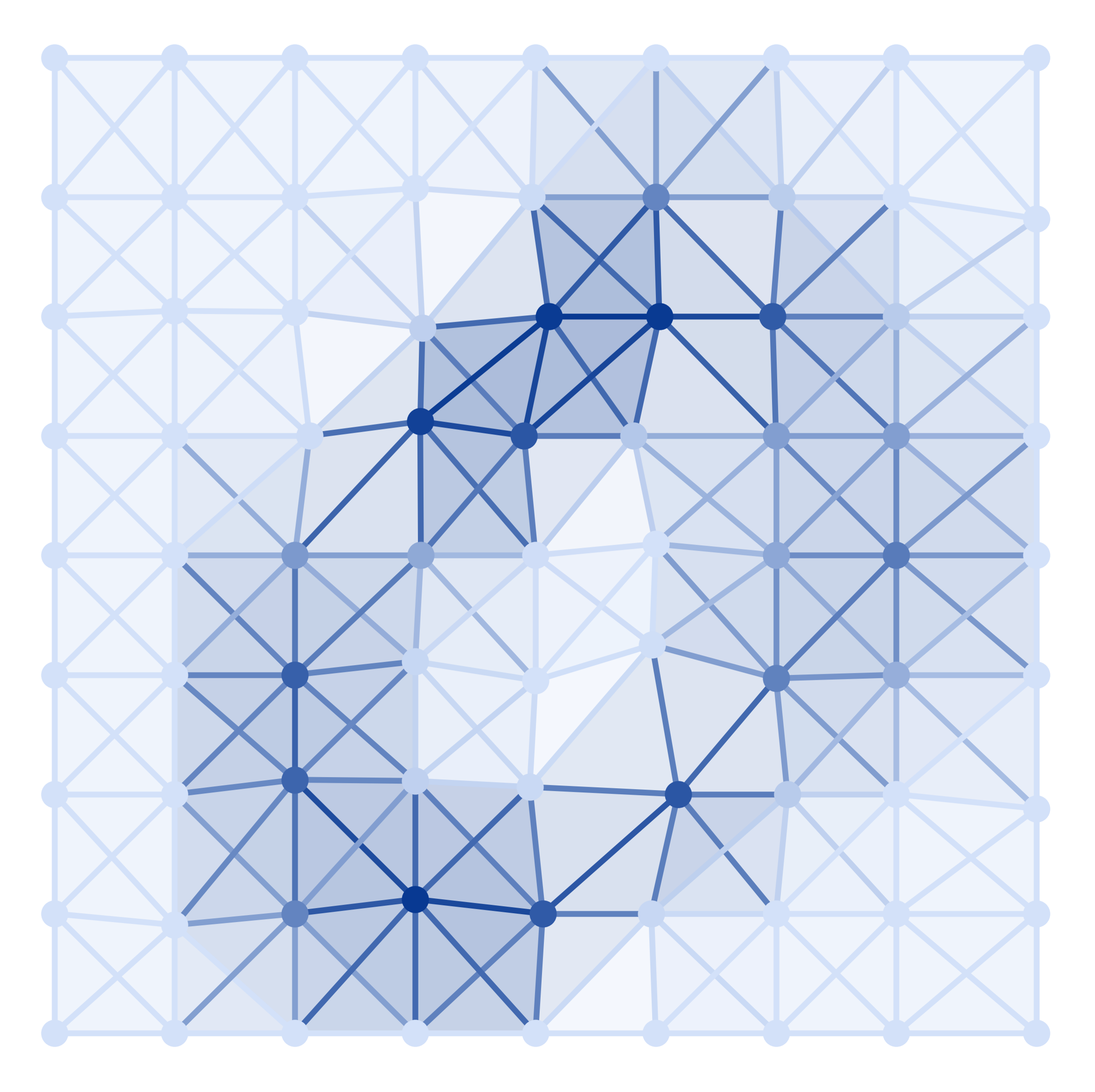}
		\centering{\small (b)}
	\end{minipage}
	\begin{minipage}[c]{.2\textwidth}
		\includegraphics[width=\textwidth]{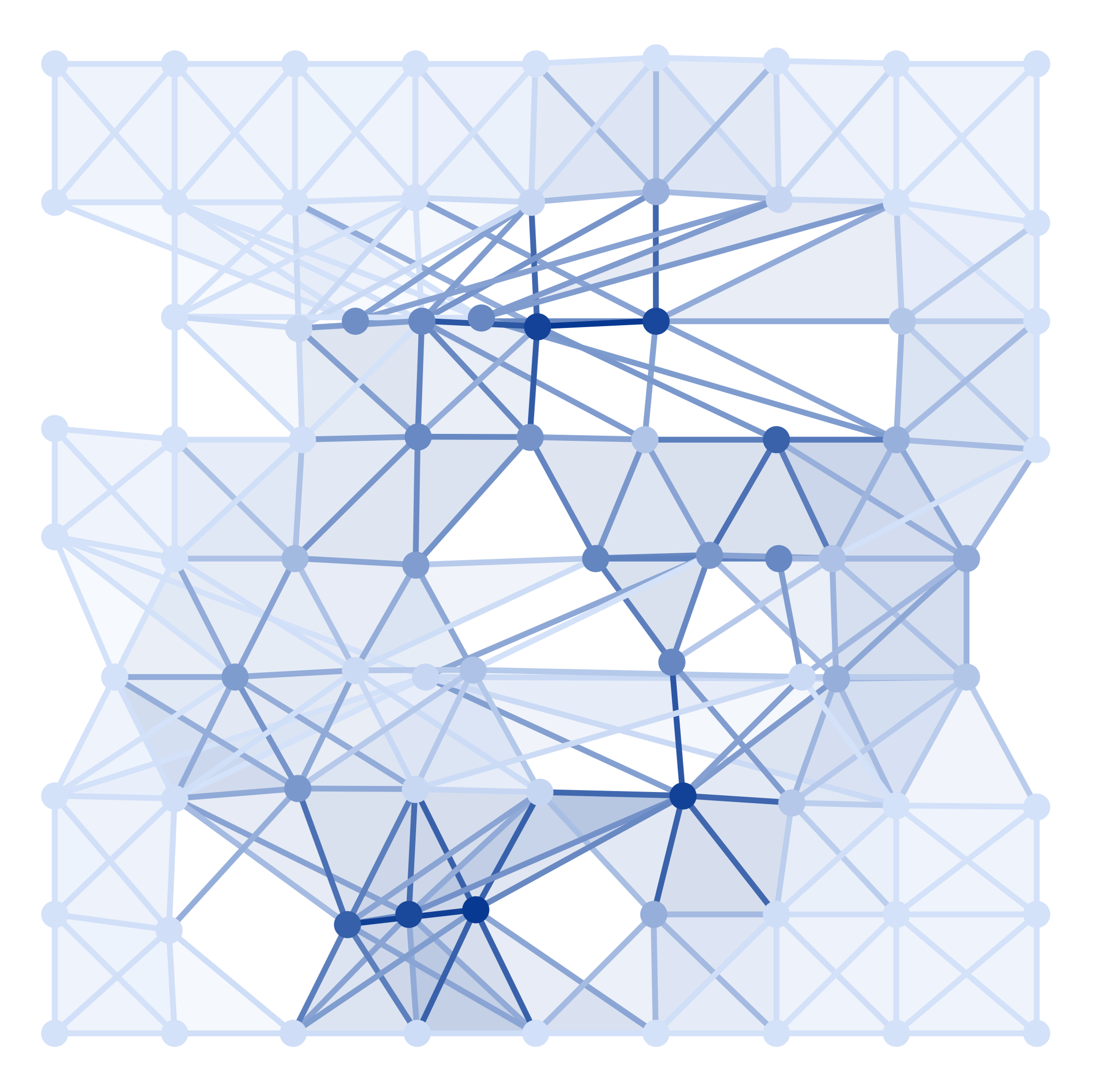}
		\centering{\small (c)}
	\end{minipage}
	\begin{minipage}[c]{.2\textwidth}
		\includegraphics[width=\textwidth]{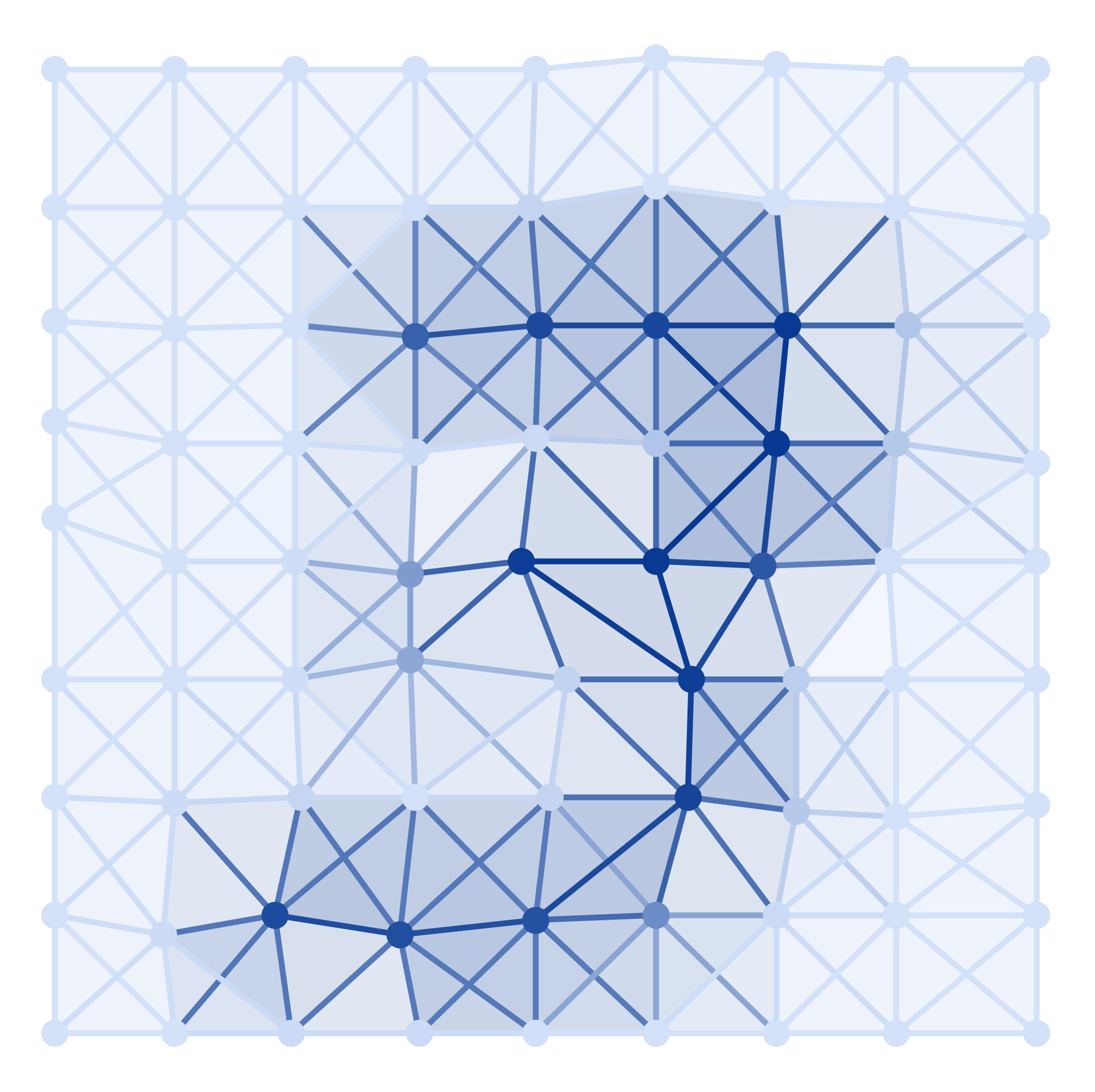}
		\centering{\small (d)}
	\end{minipage}
\caption{\small{Linear and convex clustering mixup of complexons.
(a) Clusterpath from complexons estimated from two sets of Vietoris-Rips complexes, where each complex is formed from i.i.d. points sampled from one of two shapes in $\mathbb{R}^2$, a circle and a figure eight.
As $\lambda$ increases, complexes generated from the same shape fuse together before both shapes coalesce, but we maintain knowledge of the overall spread of data, as Class 1 fuses before Class 0.
(b) Superpixel simplicial complex for MNIST digit 0. 
(c) Superpixel simplicial complex sampled from the complexon $W_\mathrm{new}=0.5\hat{W}_0 + 0.5\hat{W}_3$, where $\hat{W}_0$ and $\hat{W}_3$ denote complexons estimated from the MNIST digits 0 and 3.
(d) Superpixel simplicial complex for MNIST digit 3.
}}
\label{fig:vrsc_mixup}
\end{figure*}

\section{Preliminaries}
\label{s:bg}

\noindent\textbf{Simplicial complexes.}
A simplicial complex $K$ is a finite collection of finite sets of elements, or simplices, that are closed under restriction, that is, for every subset $\sigma\in K$, all strict subsets $\sigma'\subset \sigma$ must also be in $K$~\cite{munkresTopology2000}.
We let $K^{(d)}\subseteq K$ denote the subset of $K$ containing simplices in $K$ with cardinality $d+1$, which are said to have dimension $d$. 
The dimension of a simplicial complex $K$ is $d$, where $d$ is the largest dimension for which $K^{(d)}$ is not empty.
We may view the subset $K^{(0)}$ as the nodes in $K$, $K^{(1)}$ as edges, $K^{(2)}$ as triangles, and so on. 
We further define the degree of node $i$ at dimension $d$ of $K$ as $D^{(d)}_i:=|\{\sigma\in K^{(d)} : i\in\sigma\}|$.


For a pair of simplicial complexes $F$ and $K$, the homomorphism density of $F$ in $K$ is $t(F,K) = \mathrm{hom}(F,K)/|F^{(0)}|^{|K^{(0)}|}$, where $\mathrm{hom}(F,K)$ denotes the number of homomorphisms from $F$ to $K$~\cite{roddenberry2023limits}.
Intuitively, $t(F,K)$ represents the number of occurrences of $F$ in $K$ while preserving simplices.

\medskip

\noindent\textbf{Mixup for graph data augmentation}.
Mixup has enjoyed well-deserved popularity as an intuitive and efficient data augmentation method~\cite{zhang2018mixup}.
The classical mixup method obtains new samples as convex combinations of pairs of samples from different classes. 
Variants have been proposed for several domains and applications, including graph mixup~\cite{han2022gmixup,navarroGraphmadGraphMixup2023}, interpolation in an embedding space~\cite{verma2019manifold}, and nonlinear implementations~\cite{navarroGraphmadGraphMixup2023}.

Despite the rapid development of mixup for graphs, it remains difficult due to their non-Euclidean nature, so mixup in a continuous embedding space remains a popular approach~\cite{han2022gmixup,navarroGraphmadGraphMixup2023,wangMixupNodeGraph2021}.
However, projection from the graph domain onto a lower-dimensional space may lose critical semantic information, and the potential for information loss is even greater for higher-order networks, for which there are far fewer data augmentation methods~\cite{wei2022augmentations}.

\medskip

\noindent\textbf{Limit objects for networks}.
The increasing presence of large graphs, such as the internet, motivates the concept of graph limits.
The graphon was thus introduced as the limit of a convergent sequence of dense graphs~\cite{lovaszLargeNetworksGraph2012a}.
Simplicial complex sequences have recently received an analogous limit object known as the complexon~\cite{roddenberry2023limits}.
Formally, a complexon $W$ is a measurable function
\be
    W:\bigsqcup_{d\geq 1} [0,1]^{d+1}\rightarrow[0,1]
\ee
that is symmetric in its coordinates at every dimension.
We represent the complexon at dimension $d$ as $W^{(d)}$, and we may view a graphon as a complexon of dimension 1, $W^{(1)}$.
Similarly to graphons~\cite{lovaszLargeNetworksGraph2012a}, complexons not only represent limit objects but also can be used as a generative model to sample new simplicial complexes~\cite{roddenberry2023limits}.

We may also define the homomorphism density of the simplicial complex $F$ in the complexon $W$ as 
\be
    t(F,W)=\int_{[0,1]^{F^{(0)}}} \prod_{\sigma\in F} W(\zeta_\sigma)d\zeta,
\ee
where $\zeta_\sigma$ corresponds to indexing $(\zeta_1,\dots,\zeta_{|F^{(0)}|})\in[0,1]^{F^{(0)}}$ by $\sigma\subseteq F^{(0)}$~\cite{roddenberry2023limits}.

\section{Methodology}
\label{s:method}

Given a dataset of labeled simplicial complexes $\ccalD=\{(K_i,y_i)\}_{i=1}^T$, we aim to generate a synthetic dataset $\ccalD'=\{(K'_i,y'_i)\}_{i=1}^{T'}$ such that a classifier $f$ trained on the augmented dataset $\ccalD\cup\ccalD'$ achieves a higher accuracy predicting the labels of unseen samples compared to training solely on the original dataset $\ccalD$.
We present SC-MAD for simplicial complex data augmentation following a three-step procedure: (1) We embed the existing simplicial complexes onto a continuous space, which we select as the space of complexons~\cite{roddenberry2023limits}, (2) we perform mixup either via the efficient pairwise linear mixup~\cite{zhang2018mixup} or the more informative convex clustering mixup~\cite{navarroGraphmadGraphMixup2023}, and (3) we sample complexon mixtures from the interpolants obtained from mixup and generate new simplicial complexes from those mixtures. 

We now discuss the intuition behind the complexon as the embedding space.
Step (1) of SC-MAD is common for mixup methods, where samples are interpolated in an embedding space~\cite{verma2019manifold,han2022gmixup,navarroGraphmadGraphMixup2023}.
The choice of embedding space is adaptable to a user's desired preserved characteristics when obtaining mixtures, and the complexon is a natural choice for the continuous treatment of simplicial complexes.
First, as a Euclidean object, it enjoys amenability to interpolation for mixup.
Second, the complexon can be used as a random simplicial complex model, representing a family of simplicial complexes~\cite{roddenberry2023limits}.
For complex objects such as ours, a stochastic inversion is desirable for generating many views of simplicial complexes from the same complexon mixture.
Third, invertible embeddings permit learning in the original space, mitigating information loss from lower-dimensional projections.

\subsection{SC-MAD steps}
\label{ss:steps}

We elaborate on each step of SC-MAD in the sequel.
Of primary importance is how to convert simplicial complexes into complexons.

\medskip

\noindent
\textbf{Step (1) Complexon estimation.}
We perform complexon estimation for each labeled simplicial complex $\{(K_i,y_i)\}_{i=1}^T$ to obtain a set of complexon embeddings $\{(\hat{W}_i,y_i)\}_{i=1}^T$.
The task of estimating a graphon from a single graph is well studied, for which there are several computationally efficient and effective methods~\cite{chan2014consistent,bickelNonparametricViewNetwork2009,yangNonparametricEstimationTesting2014}.
We adapt sorting-and-smoothing (SAS) for \emph{graphon} estimation ~\cite{chan2014consistent} to \emph{complexon} estimation, where SAS consists of (1) sorting nodes by degree and (2) estimating edge probability by computing network histograms.
Inspired by this, we obtain node orderings at every dimension and jointly apply them to sort nodes with more information than if we were to only sort by the number of edges as with graphons.

We first sort nodes in a given $d$-dimensional simplicial complex $K$ with $N$ nodes by computing the following sum
\begin{equation}\label{eq:degsum}
    D_i = \sum_{c=1}^{d} \tau^c D_i^{(c)}
\end{equation}
for every node $i\in\{1,2,\dots,N\}$, where $\tau\in(0,1)$ and $D^{(c)}_i$ is the degree of node $i$ at dimension $c$ as in Section~\ref{s:bg}.
Reordering the nodes in $K$ by the degree sum in \eqref{eq:degsum} gives the sorted simplicial complex $K_\phi$.
We obtain a piecewise constant complexon $\hat{W}$ as a simplicial complex histogram, whose values at dimension $c$ measure the frequencies of $c$-simplices of $K_\phi$ in histogram bins~\cite{chan2014consistent}.
More specifically, for any $\zeta^{(c)}=(\zeta_1,\dots,\zeta_{c+1})\in[0,1]^{c+1}$, we obtain
\begin{alignat}{3}
    &\hat{W}^{(c)}_\circ\left( \zeta^{(c)} \right) = 
    \frac{1}{h^{c+1}}
    \sum_{j_1=1}^{h} \cdots 
    &\nonumber\\
    & \qquad
    \sum_{j_{c+1}=1}^h
    \mathbb{I}\left\{
    (q(\zeta_1) h+j_1,\dots,q(\zeta_{c+1}) h+j_{c+1})
    \in K_\phi
    \right\}, \label{eq:sas_pt1}
    &
\end{alignat}
where $h>0$ denotes the number of nodes in each bin and we let $q(\zeta) = \max \{\ceil{\zeta \floor{N/h}}-1,0\}$.
The estimate $\hat{W}_\circ$ approximates the \emph{faceted} complexon~\cite{roddenberry2023limits}. 
Hence, we obtain the final complexon estimate $\hat{W}$ by computing 
\begin{equation}\label{eq:sas_pt2}
    \hat{W}^{(c)}(\zeta^{(c)}) = \hat{W}^{(c)}_\circ(\zeta^{(c)}) 
    \left(
    \prod_{\zeta \subset \zeta^{(c)}} \hat{W}^{(|\zeta|-1)}_\circ(\zeta)
    \right)^{-1},
\end{equation}
for every $\zeta^{(c)}=(\zeta_1,\dots,\zeta_{c+1})\in[0,1]^{c+1}$.
The complexon estimation in \eqref{eq:sas_pt1} and \eqref{eq:sas_pt2} generalizes the popular SAS graphon estimation while accounting for interactions across dimensions for higher-order objects.

\medskip

\noindent\textbf{Step (2) Complexon mixup.}
Once the labeled complexon estimates $\{(\hat{W}_i,y_i)\}_{i=1}^T$ are obtained, we can then apply linear or convex clustering mixup. 
For pairwise linear mixup, we select a pair of complexons $\hat{W}_i$ and $\hat{W}_j$ such that $y_i\neq y_j$ and interpolate as 
\begin{equation}\label{eq:linmixup}
    W_\mathrm{new} = (1-\lambda)\hat{W}_i + \lambda\hat{W}_j,
\end{equation}
where $\lambda\in[0,1]$.
For convex clustering mixup, we solve the following optimization problem~\cite{pelckmans2005convex,hocking2011clusterpath}
\begin{alignat}{3}
\hat{U}(\lambda) = \argmin_{ U } 
\sum_{i=1}^T \rho_{\mathrm{fid}}(U_i,\hat{W}_i) 
+ \frac{\lambda}{1-\lambda}\sum_{i<j} w_{ij}\rho_{\mathrm{fus}}(U_i,U_j), \label{eq:cvxclust}
\end{alignat}
where $\lambda\in[0,1)$ is the tunable mixup parameter, $w_{ij}\geq 0$ is the weight determining the level of fusion between $\hat{W}_i$ and $\hat{W}_j$, and the functions $\rho_\mathrm{fid}$ and $\rho_\mathrm{fus}$ respectively quantify fidelity and fusion~\cite{navarroGraphmadGraphMixup2023}.
We choose the following convex functions 
\begin{alignat}{3}&    \rho_\mathrm{fid}(W_1,W_2)=
    \textstyle\sum_{c=0}^{d} \int_{[0,1]^{c+1}} (W^{(c)}_1(\zeta)-W^{(c)}_2(\zeta))^2d\zeta,
    &\nonumber\\&
    \rho_\mathrm{fus}(W_1,W_2)=
    \textstyle\sum_{c=0}^{d} \int_{[0,1]^{c+1}} |W^{(c)}_1(\zeta)-W^{(c)}_2(\zeta)|d\zeta. \nonumber
&\end{alignat}

The clusterpath $\hat{U}(\lambda)=\{\hat{U}_i(\lambda)\}_{i=1}^T$ returns complexon mixtures at each $\lambda\in[0,1]$, with $\hat{U}(1)=\{\frac{1}{T}\sum_j \hat{W}_j\}_{i=1}^T$ by definition.
When $\hat{U}_i(\lambda)=\hat{U}_j(\lambda)$, we say that the value is the mixture of $\hat{W}_i$ and $\hat{W}_j$, where $\hat{W}_i$ and $\hat{W}_j$ are fused.
The mixup parameter $\lambda$ determines how similar to the original complexons $\hat{W}$ the mixtures should be.
When $\lambda=0$, $\hat{U}(0)=\{\hat{W}_i\}_{i=1}^T$ returns the original complexons, and as $\lambda$ increases, complexons begin to fuse into clusters.
We encourage the clusterpath $\hat{U}(\lambda)$ to identify class differences for downstream classification by letting $w_{ij}=1$ when $y_i=y_j$ and $w_{ij}=\epsilon$ otherwise for some $\epsilon>0$.
For further implementation details, we refer the reader to~\cite{navarroGraphmadGraphMixup2023}.
Once we obtain the clusterpath $\hat{U}(\lambda)$ from \eqref{eq:cvxclust}, we select complexon mixtures $W_\mathrm{new}=\hat{U}_i(\lambda)$ by choosing $\lambda\in[0,1]$ and $i\in\{1,2,\dots,T\}$.
A visualization of the clusterpath $\hat{U}(\lambda)$ for two sets of Vietoris-Rips complexes is shown in Fig.~\ref{fig:vrsc_mixup}a.

\medskip

\noindent\textbf{Step (3) Simplicial complex sampling.}
As with graphons, there is an analogous process for sampling simplicial complexes from complexons~\cite{roddenberry2023limits}.
Given a set of nodes $K^{(0)}_\mathrm{new}$, we sample edges from the complexon $W_\mathrm{new}$ as
\begin{subequations}
\begin{alignat}{3}&
    \zeta_i\sim\mathrm{Unif}([0,1])
    &&\forall~i\in K^{(0)}_\mathrm{new},
    &\nonumber\\&
    \mathbb{P}\left[(i,j)\in K_\mathrm{new}^{(1)}\right]=W_\mathrm{new}^{(1)}(\zeta_i,\zeta_j)
    &\qquad&\forall~(i,j)\in K^{(0)}_\mathrm{new}\times K^{(0)}_\mathrm{new}
    , \nonumber
&\end{alignat}
\end{subequations}
identical to that of graphons.
Beyond edges, to retain closure under restriction, we must preclude simplices whose proper subsets are not all already present in the sampled simplicial complex.
At dimension $d>1$, we add a $d$-simplex $\sigma$ to $K_\mathrm{new}^{(d)}$ with probability 
\be
    \mathbb{P}\left[
    \sigma\in K^{(d)}_\mathrm{new}
    \right]=W^{(d)}_\mathrm{new}(\zeta_{\sigma})
    \prod_{\sigma'\subset\sigma}
    \mathbb{I}\left\{ \sigma'\in K_\mathrm{new} \right\},
\ee
where $W_\mathrm{new}^{(d)}(\zeta_\sigma)$ represents the probability of $\sigma\in K_\mathrm{new}$ conditioned on the existence of all its proper subsets in $K_\mathrm{new}$.
Once a desired dimension is reached, the result is a simplicial complex $K_\mathrm{new}$ satisfying closure under restriction.
Further details are provided in~\cite{roddenberry2023limits}.
We can then sample any number of new simplicial complexes from one complexon $W_\mathrm{new}$, generating multiple views from the same model whose structural characteristics are preserved.

\subsection{Class structure in complexon mixtures}

Mixup aims to generate new samples with characteristics from multiple classes.
We theoretically show that the complexon mixtures $W_\mathrm{new}$ from linear mixup~\eqref{eq:linmixup} or convex clustering mixup \eqref{eq:cvxclust} contain a mixture of class-dependent structural characteristics from multiple simplicial complexes.
In particular, we assume that for each class $y$, there is a finite set of discriminative simplicial complexes $\ccalF_y$ such that for every labeled simplicial complex $(K,y)$, there exists at least one $F\in\ccalF_y$ that is a subcomplex of $K$~\cite{han2022gmixup}, that is, there is a homomorphism from $F$ to $K$.
We present the following result on the structural similarities between a complexon mixture and one of the complexons, inspired by a similar result for graphon mixup~\cite{han2022gmixup}.
\begin{mytheorem}\label{thm:homdens}
    Consider a set of simplicial complexes $\{(K_i,y_i)\}_{i=1}^T$ from which we estimate a set of complexons $\{\hat{W}_i\}_{i=1}^T$.
    Let the convex combination $W_\mathrm{new}=\sum_{i=1}^T \gamma_i \hat{W}_i$ for $\sum_{i=1}^T\gamma_i=1$ denote a complexon mixture from \eqref{eq:linmixup} or \eqref{eq:cvxclust}, and let $\ccalF_{y_j}$ be the discriminative simplicial complex set for class $y_j$.
    For any $F\in\ccalF_{y_j}$, we present the following upper bound on the homomorphism density difference for the complexon mixture $W_\mathrm{new}$ and the estimate $\hat{W}_j$
    \begin{equation}\label{eq:homdens}
        |t(F,W_\mathrm{new})-t(F,\hat{W}_j)| \leq
        \sum_{i\neq j} \gamma_i
        \rho_\square (\hat{W}_i,\hat{W}_j;(\beta^{(c)})_{c\geq 1} ),
    \end{equation}
    where $\beta^{(c)} = |F^{(c)}|$ is the number of $c$-simplices in $F$, and $\rho_\square$ denotes the cut distance for complexons as described in~\cite{roddenberry2023limits}.
\end{mytheorem}
\medskip

\noindent\textbf{Proof sketch.}
We omit a full proof of Theorem~\ref{thm:homdens} for space and provide a brief description instead.
We rely on the Counting Lemma for Complexons~\cite{roddenberry2023limits} to bound the homomorphism density difference for a simplicial complex in two different complexons. Then, by Jensen's inequality, we separate each term of the complexon mixture to obtain the sum on the right-hand side of \eqref{eq:homdens}.
$\hfill\blacksquare$

\medskip

Note that for complexons of dimension 1, when $\gamma_i=\lambda$, $\gamma_j=1-\lambda$, and $\gamma_k=0$ for every $k\neq i,j$, Theorem~\ref{thm:homdens} reduces to the result for pairwise graphon mixup in~\cite{han2022gmixup}.
Our result generalizes that of~\cite{han2022gmixup} by allowing arbitrary convex combinations and any complexon dimension.
Theorem~\ref{thm:homdens} shows that the discriminative structure of a given class $y_j$ is present in the mixture $W_\mathrm{new}$, and this presence increases as $\hat{W}_j$ grows closer to the remaining complexons in the set.
Furthermore, since convex clustering obtains mixtures of every complexon in a set, the complexon mixtures obtained from \eqref{eq:cvxclust} will contain the discriminative structure for every class.

\begin{table}[t]
    \footnotesize
    \centering
    \begin{tabular}{m{4.7em} m{5.2em} m{6em} m{6em}}
    \hline
        \multicolumn{2}{c}{\bf Method} &
        \multicolumn{1}{c}{\bf Vietoris-Rips} & 
        \multicolumn{1}{c}{\bf MNIST} \\
    \hline
        Data mixup & Label mixup & \\
    \hhline{|-|-|}
        None & None & $0.631\pm0.167$ & $0.782\pm0.051$ \\ 
        \hline
        \multirow{4}{*}{Linear} & Linear & $0.709\pm0.051$ & $0.802\pm0.111$ \\
         & Sigmoid & $\mathbf{0.719\pm0.084}$ & $0.687\pm0.088$ \\
         & Logit & $0.594\pm0.146$ & $0.705\pm0.033$ \\
         & Cvx. clust. & $0.669\pm0.193$ & $0.805\pm0.057$ \\ \hline
        \multirow{4}{*}{Cvx. clust.} & Linear & $0.688\pm0.196$ & $0.804\pm0.110$ \\
         & Sigmoid & $0.688\pm0.156$ & $\mathbf{0.819\pm0.072}$ \\
         & Logit & $0.709\pm0.064$ & $0.817\pm0.049$ \\
         & Cvx. clust. & $\mathbf{0.738\pm0.057}$ & $\mathbf{0.856\pm0.052}$ \\
    \hline
    \end{tabular}
    \caption{\small{Simplicial complex classification accuracy. The top performing methods are \textbf{bolded}.}}
    \label{t:results}
\end{table}

\section{Numerical evaluation}
\label{s:results}

We evaluate SC-MAD for generating labeled simplicial complexes to improve classification accuracy.
We use a simplicial convolutional network (SCN) as the architecture for each of the following simulations~\cite{ebliSimplicialNeuralNetworks2020}, and we compare model prediction performance with and without data augmentation.
We perform simplicial complex mixup via linear mixup~\eqref{eq:linmixup}, denoted ``Linear'', and convex clustering mixup~\eqref{eq:cvxclust}, denoted ``Cvx. clust.'', as described in Section~\ref{s:method}.
For both methods, we let $\lambda\sim\mathrm{Unif}([0,1])$.
We also compare four methods for mixup of labels~\cite{navarroGraphmadGraphMixup2023}.
We interpolate labels $y_i$ and $y_j$ given the mapping $g:[0,1]\rightarrow[0,1]$ as
\be
    y_\mathrm{new} = \left(1-g(\lambda)\right)y_i + g(\lambda)y_j.
\ee
For $a>0$, we consider ``Linear'' mixup $g(\lambda)=\lambda$; ``Sigmoid'' mixup $g(\lambda)=1/(1+\exp\{-a(2\lambda-1)\})$; ``Logit'' mixup $g(\lambda)=\log(\lambda/(1-\lambda))/2a+1/2$; and ``Cvx. clust.'', convex clustering label mixup as introduced in~\cite{navarroGraphmadGraphMixup2023}.

\medskip

\noindent\textbf{Synthetic data.}
Consider two classes of Vietoris-Rips complexes, where each complex is formed from i.i.d. points sampled from one of two shapes in $\mathbb{R}^2$, a circle and a figure eight.
We perform simplicial complex classification to identify from which shape each complex is sampled.
We present the shape classification accuracy for each method in the column of Table~\ref{t:results} denoted ``Vietoris-Rips''. 
The first row of Table~\ref{t:results} corresponds to the original dataset with no data augmentation. 
The column ``Data mixup'' indicates the simplicial complex mixup method and ``Label mixup'' the label mixup method. 

In all cases but one, data augmentation via mixup improves prediction performance.
We observe the greatest increase in classification accuracy when using convex clustering for both data and labels, as expected due to the more informative sampling of new labeled simplicial complexes.
We emphasize the practicality of convex clustering for mixup as we achieve superior performance without requiring a specified mixup function for data or labels, nor do we require a user-defined sampling mechanism for the mixup parameter $\lambda$~\cite{zhang2018mixup}.
We thus demonstrate the viability of the complexon for interpolating in the higher-order simplicial complex space.
With this choice of interpolation space, we reap the advantages of mixup for improving performance even for such complex structures.

\medskip

\noindent\textbf{Image data.}
We also evaluate our proposed mixup on the MNIST image dataset~\cite{lecun2010mnist}.
Any image can be represented as a superpixel graph, where each node corresponds to a cluster of pixels denoting meaningful regions and each edge connects nodes that are adjacent in pixel space~\cite{gohSimplicialAttentionNetworks2022a}.
To encode richer visual information, we add triangles for every clique of three nodes in the superpixel graph, resulting in a set of simplicial complexes modeling related regions within each image.
In Fig.~\ref{fig:vrsc_mixup}b and d, we show simplicial complex representations of two handwritten digits in the MNIST dataset, while in Fig.~\ref{fig:vrsc_mixup}c, we present the simplicial complex sampled from a complexon mixture obtained via linear mixup of the original two images with $\lambda=0.5$.
We obtain a mixed superpixel simplicial complex that exhibits structural interpolation rather than mere pixel-wise value mean.
In particular, the mixture in Fig.~\ref{fig:vrsc_mixup}c not only mixes pixel values by interpolating simplex features but also changes how image regions, represented by nodes, are connected, modifying which regions are relevant to which.

A comparison of our results for simplicial complex classification on a subset of three classes of MNIST images is shown in the column of Table~\ref{t:results} denoted ``MNIST''.
As image classification is well understood, superpixel network classification serves as a useful benchmark for comparing simplicial complex-based learning methods.
Convex clustering for both images and labels results in the greatest increase in classification accuracy over the original superpixel dataset.
This demonstrates the power of convex clustering for providing informative synthetic samples for real-world multiclass data.
Moreover, almost all mixup methods achieve superior performance relative to the original dataset, including those that apply different methods for label and image mixup.
This motivates future investigation in pursuing optimal ways to mixup data and labels.

\section{Conclusion}
\label{s:conclusion}

In this work, we presented simplicial complex mixup via complexons, the limit object of convergent simplicial complex sequences.
With the continuous complexon, we were able to exploit the efficiency of linear pairwise mixup along with the effectiveness of convex clustering mixup for discrete, irregular simplicial complexes.
The success of our method for simplicial complexes implies the practicality of exploring limit objects for other data types to perform useful tasks typically limited to Euclidean data, without needing domain expertise or computationally intensive approaches.
Furthermore, we theoretically validated our ability to manipulate simplicial complexes while preserving structural characteristics, so the ubiquitous use of graphs in many applications can be naturally extended to simplicial complexes.
We may more easily adopt these higher-order networks for other useful fields that graphs already occupy, such as social network analysis.

\bibliographystyle{ieeetr}
\bibliography{citations}

\end{document}